\documentclass[runningheads]{llncs}
\usepackage[T1]{fontenc}
\usepackage{graphicx}
\usepackage[x11names, svgnames, rgb]{xcolor}
\usepackage[utf8]{inputenc}
\usepackage{tikz}
\usetikzlibrary{snakes,arrows,shapes}
\usepackage{amsmath}
\usepackage{amsfonts}
\usepackage{amssymb}

\newcommand{\x}{\mathbf{x}}
\renewcommand{\u}{\mathbf{u}}

\renewcommand{\L}{\mathcal{L}}

\renewcommand{\a}{\pmb{\alpha}}
\renewcommand{\b}{\pmb{\beta}}
\newcommand{\inv}[1]{\frac{1}{#1}}

\newcommand{\abs}[1]{|#1|}

\newcommand{\N}{\mathcal{N}}
\newcommand{\params}{\mathbf{\Theta}}

\DeclareMathOperator*{\argmax}{arg\,max}

\begin{document}
{
    \centering
    \color{lightgray} \small
    This is a pre-print. \\
    \vspace{-0.5ex}
    Aside from formatting differences, it is identical to the version published \\
    \vspace{-0.5ex}
    in Springer Lecture Notes for Computer Science Vol. 13886 \\
}
\title{
    Out-of-Distribution Detection for Adaptive Computer Vision
}
%
%
\author{
    Simon Kristoffersson Lind\inst{1} --
    \texttt{simon.kristoffersson\_lind@cs.lth.se} \and \\
    Rudolph Triebel\inst{3,4} --
    \texttt{rudolph.triebel@dlr.de} \and \\
    Luigi Nardi\inst{1,2} --
    \texttt{luigi.nardi@cs.lth.se} \and \\
    Volker Krueger\inst{1} --
    \texttt{volker.krueger@cs.lth.se}
}
\authorrunning{S. Kristoffersson Lind et al.}
%
\institute{
    Lund University LTH \and
    Stanford University \and
    German Aerospace Center DLR \and
    Technical University of Munich
}

{\let\newpage\relax\maketitle}

\subsubsection*{Corresponding Author:} Simon Kristoffersson Lind

\begin{abstract}
It is well known that computer vision can be unreliable when faced with previously unseen imaging conditions.
This paper proposes a method to adapt camera parameters according to a normalizing flow-based out-of-distibution detector.
A small-scale study is conducted which shows that adapting camera parameters according to this out-of-distibution detector
leads to an average increase of 3 to 4 percentage points in mAP, mAR and F1 performance metrics of a YOLOv4 object detector.
As a secondary result, this paper also shows that it is possible to train a normalizing flow model for out-of-distribution detection
on the COCO dataset, which is larger and more diverse than most benchmarks for out-of-distibution detectors.

\end{abstract}

\section{Introduction}
In the past decade computer vision has become a de facto component in autonomous systems.
However, it is well known that vision can be unreliable when faced with new,
previously unseen situations.
Uncertainty is especially abundant in the field of robotics,
where the autonomous agents are expected to perform tasks in real-world scenarios that often differ significantly from any training data.

There is an abundance of research to suggest that \textit{Out-of-Distribution} (OOD) detection
can help improve the reliability of vision algorithms \cite{SurveyUncertainty}.
Relevant literature reasons that OOD detection allows systems to take action when uncertainty is encountered \cite{SurveyUncertainty}.
However, there is little research to show that OOD detection helps the reliability of autonomous systems in practice.
Most literature on OOD detection tends to focus either on synthetic benchmarks,
or controlled real-world scenarios where in- and out-of-distribution examples are very distinct.
Neither of these scenarios are reflective of the diverse situations that may be encountered by
autonomous systems in practice \cite{SurveyUncertainty}.

OOD detectors are commonly used by setting a threshold OOD score \cite{kirichenko},
and simply discarding any input that falls beyond this threshold.
While this makes intuitive sense, on the basis that vision systems tend to be unreliable when faced with OOD data,
it only utilises one end of the OOD scores.
We hypothesize that a machine learning model will on average perform more reliably when an input has a lower OOD score.
Therefore we propose to use an OOD detector as a quality metric, or un-normalized confidence;
the lower the OOD score, the more certain we can be about our vision system's output.

\subsection{Our Contributions}
\begin{itemize}
    \item We expand the current knowledge on OOD detection by showing that it is possible to train
          normalizing flow models for OOD detection on large, diverse datasets while getting sensible results.
    \item We introduce a novel method for utilising OOD detection as a quality metric in autonomous systems
          and show that it can lead to reliability improvements for vision tasks.
\end{itemize}

\section{Related Work}
There has been some previous work investigating the uses of OOD detection in autonomous systems.
For example Wellhausen \textit{et al}. \cite{wellhausen2020} perform anomaly detection on image data
collected from a robot in real-world terrain.
However, their main purprose is the evaluation of different OOD detection algorithms,
and not an application of OOD detection to real-world operation.

Yuhas \textit{et al}. \cite{yuhas2021} evaluate OOD detection as an emergency breaking system for an autonomous car,
though not in real-world operation, but on a custom test track.

McAllister \textit{et al}. \cite{mcallister2019} perform real-world crash avoidance experiments with an autonomous car.
However, their method does not perform OOD detection directly.
Instead they use a variational autoencoder to generate in-distribution samples from an out-of-distribution image,
which they use as a measure of uncertainty.

Additionally, there is an abundance of work on OOD detection where the authors have taken care to
ensure that in- and out-distributions are disjoint, for example \cite{Deecke2019,Hsu2020,Liang2017,Winkens2020,Ren2019,Mohseni2020}.
These works either use entirely different datasets as in- and out-distributions,
or separate classes as in- and out-distributions.
It is our belief that neither of these scenarios are reflective of real-world operation.

\section{Background}
\subsection{Out-of-Distribution Detection}
Conceptually, out-of-distibution detection is simple.
First an in-distribution is defined, for example the training data for a machine learning-based vision system.
Then, everything that deviates from said in-distribution is said to be out-of-distribution \cite{OODdetection}.

More formally,
we assume that our vision system operates on data that is sampled I.I.D from some distribution $P_{train}(\x)$.
In practice, however, the system may encounter data that comes from a different distribution,
in which case the system will produce unreliable results.
Therefore, it is desirable to detect when data lies outside of the original distribution $P_{train}(\x)$.

\subsection{Normalizing Flows}
This section aims to provide a basic introduction to normalizing flows.
For more details we refer to Papamakarios \textit{et al}. \cite{papamakarios}.

Normalizing flows have emerged in the last decade as a way to model complex probability distributions \cite{papamakarios}.
Let us assume that we want to evaluate or sample from a distribution $p_x(\x)$, $\x \in \mathbb{R}^D$,
and that $p_x$ is intractable to evaluate or sample from.
A normalizing flow can allow us to evaluate or sample from $p_x$
by transforming $p_x(\x)$ into a tractable distribution $p_u(\u)$, $\u \in \mathbb{R}^D$.

In order to model $p_x(\x)$, an invertible transformation $T$ is constructed such that
\[
    \x = T(\u), \ \ \u = T^{-1}(\x)
\]
where $\u \sim p_u(\u)$.
Here, we are free to define $p_u(\u)$, for example $\u \sim \N(\mathbf{0}, \mathbf{1})$.

By additionally requiring both $T$ and $T^{-1}$ to be differentiable,
it is possible to recover $p_x(\x)$ from $p_u(\u)$ \cite{rudin2006,bogachev2007,papamakarios}:
\begin{equation} \label{eq:px}
    p_x(\x) = p_u(\u) \abs{\det J_T(\u)}^{-1} = p_u(T^{-1}(\x)) \abs{\det J_{T^-1}(\x)}
\end{equation}
where $J_T$ and $J_{T^{-1}}$ are the Jacobians of $T$ and $T^{-1}$ respectively.

This type of transformation is called a \textit{diffeomorphism}.
It has been shown that it is always possible to construct a diffeomorphism $T$
under reasonable assumptions about $p_x(\x)$ and $p_u(\u)$ \cite{papamakarios}.
However, analytically constructing $T$ is often infeasible in practice.
Therefore, it is desirable to learn an approximation of $T$.
Henceforth we will denote $T(\x; \params)$ as an approximation of $T$ with learnable parameters $\params$.

An important property is that the composition of two diffeomorphisms
$T_2 \circ T_1$ is also a diffeomorphism \cite{papamakarios}:
\[
    (T_2 \circ T_1)^{-1} = T_1^{-1} \circ T_2^{-1}
\]
\[
    \det J_{T_2 \circ T_1} (\u) = \det J_{T_2}(T_1(\u)) \cdot \det J_{T_1}(\u)
\]
This property opens up for many ways to construct powerful approximations $T(\x; \params)$.
In this work, we focus on a particular model of so-called \textit{affine coupling layers} \cite{papamakarios}.

Affine coupling layers first split the input $\x$ into two parts \cite{papamakarios}:
\[
    \x = [x_i \ | \ i \in \mathcal{I}], \ \ \mathcal{I} = 1,2,\ldots,D
\]
\[
    \x_1 = [x_i \ | \ i \in \mathcal{A}], \ \ \x_2 = [x_i \ | \ i \in \mathcal{B}],
    \ \
    \mathcal{A} \cup \mathcal{B} = \mathcal{I},
    \ \
    \mathcal{A} \cap \mathcal{B} = \emptyset .
\]
While $\mathcal{A}$ and $\mathcal{B}$ can be chosen arbitrarily,
the most common choice is to simply split $\x$ into two halves:
\[
    \mathcal{A} = 1,2,\ldots,\frac{D}{2}, \ \
    \mathcal{B} = \frac{D}{2}+1,\frac{D}{2}+2,\ldots,D .
\]
Then, $\x_1$ is transformed as a function of $\x_2$, while $\x_2$ is left as-is:
\begin{gather*}
    \x_1' = [x_i' | i \in \mathcal{A}] = \a \cdot \x_1 + \b, \\
    \text{where} \ \ \a = \exp(F_1(\x_2; \params_1)), \ \ \b = F_2(\x_2, \params_2) .
\end{gather*}
Here, $\cdot$ denotes element-wise multiplication, and $F_1, F_2$ are arbitrary functions, for example neural networks.
The output of the coupling layer is then simply the reconstruction of $\x$ from $\x_1'$ and $\x_2$:
\begin{gather*}
    T_{coupling}(\x; \params_1, \params_2) = \x' = [x_i' | i \in \mathcal{I}], \ \
    \text{where} \ \ x_i' = x_i \forall i \in \mathcal{B}
\end{gather*}
It is trivial to see that this construction results in an invertible and differential transformation,
since $\x_1$ is subject to an affine transformation and $\x_2$ remains the same.
Coupling layers also offer the benefit of a simple Jacobian \cite{papamakarios}:
\[
    J_{T_{coupling}} = \mathbf{1} F_1(\x_2; \params_1)
\]
where $\mathbf{1}$ denotes a row vector of ones -- in other words, summing the elements of $F_1(\x_2; \params_1)$.

A common way to construct powerful learnable transformations is to simply compose several of these coupling layers,
while changing which parts of $\x$ are modified \cite{papamakarios}.
This approach is used in for example RealNVP \cite{RealNVP}, and it is what we use in our experiments.

\subsection{Normalizing Flows for Image Out-of-Distribution Detection}
Formulating OOD detection using normalizing flows is simple since we can directly compute the likelihood using (\ref{eq:px}),
which represents the belief that an example $\x$ is in-distribution \cite{kirichenko}.
Note that a high value for $p_x(\x)$ equals a low likelihood of being OOD.
It is otherwise common for OOD detection systems to use the opposite formulation,
where a high value equals high likelihood of being OOD.

Training a normalizing flow for OOD detection is done by minimizing the KL-divergence between $p_x(\x)$ and $p_u(\u)$,
which results in the following loss function \cite{papamakarios}:
\[
    \L = -\inv{N} \sum_{n=1}^N \log p_u(T^{-1}(\x_n)) + \log \abs{\det J_{T^{-1}}(\x_n)} + C
\]
where $C$ is a constant.

Since we only need the inverse transformation $T^{-1}$ for OOD detection,
we simply let $T^{-1}$ denote the forward pass of our affine coupling layers.

Despite seeming like perfect candidates for OOD detecion,
since equation (\ref{eq:px}) allows for the direct computation of the approximate probability distribution,
normalizing flows often fail in the OOD detection setting \cite{kirichenko}.
Kirichenko \textit{et al}. \cite{kirichenko} suggest that common image flow models, such as RealNVP\cite{RealNVP},
simply learn pixel correspondences,
and as such fail to grasp any semantic information that makes up the distribution.
Instead of training a flow model on images directly,
Kirichenko \textit{et al}. suggest training a flow model on features extracted from another pretrained image network.

\section{Our Method} \label{sec:method}
Based on our hypothesis that a lower OOD score would lead to better reliability,
we propose to let a vision system adapt, instead of simply discarding OOD data.
In order to bring this into practice, we create a framework for adaption around an existing vision task,
namely object detection with a YOLOv4 \cite{YOLOv4} network.

Most commercially available cameras expose a number of parameters that can alter the visual appearance of a captured image.
Common examples are: saturation, contrast, and exposure.
These camera parameters are a natural candidate for adapting to various imaging conditions.

\begin{figure}[h!]
    \includegraphics[width=0.5\textwidth]{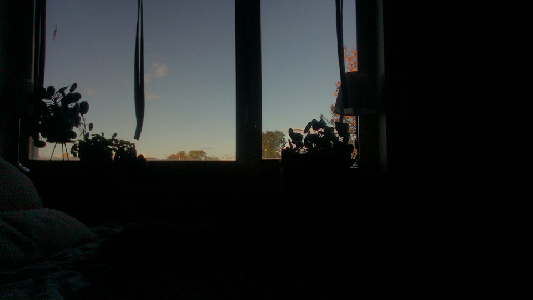}
    \includegraphics[width=0.5\textwidth]{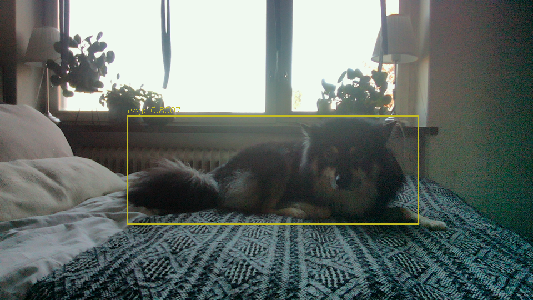}
    \caption{
        Left: camera's default parameters, auto-exposure and auto-white-balance enabled.
        Right: hand-tuned camera parameters.
    }
    \label{fig:dog}
\end{figure}

Our proposed method is to simply adapt camera parameters in order to minimize the OOD score.
Intuitively, our method can be deployed in two ways:
\begin{itemize}
    \item either: if an image is marked as OOD, adapt camera parameters to minimize OOD score,
    \item or: continuously adapt camera parameters to minimize OOD score.
\end{itemize}
While most cameras have some built-in measures to adapt, for example auto-exposure,
these are often not sufficient.
Figure \ref{fig:dog} shows two images of the same scene to illustrate this.
In one picture, the camera's default settings fail to produce an image suitable for object detection.
With hand-tuned parameters however, a more suitable image is acquired.
Note also the yellow bounding box indicates that YOLOv4 can detect the dog in the hand-tuned image,
which is not possible with the camera's default settings.

\section{Experiments and Results}
In this section we outline our two main experiments.
Section \ref{sec:model} outlines the normalizing flow model used in our experiments.
For our first experiment, in section \ref{sec:coco_flow},
we train and validate a normalizing flow model as an OOD detector using the COCO dataset \cite{COCO} as our in-distribution.
Then, in section \ref{sec:experiment},
we take the model from section \ref{sec:coco_flow} and apply it to a real-world experiment
testing our proposed method from section \ref{sec:method}.

\subsection{Normalizing Flow Model} \label{sec:model}
\begin{figure}[h!]
    \enlargethispage{100cm}
\begin{tikzpicture}[>=latex',line join=bevel,]
\begin{scope}
  \pgfsetstrokecolor{black}
  \definecolor{strokecol}{rgb}{1.0,1.0,1.0};
  \pgfsetstrokecolor{strokecol}
  \definecolor{fillcol}{rgb}{1.0,1.0,1.0};
  \pgfsetfillcolor{fillcol}
  \filldraw (0.0bp,0.0bp) -- (0.0bp,90.0bp) -- (506.0bp,90.0bp) -- (506.0bp,0.0bp) -- cycle;
\end{scope}
  \node (s) at (260.0bp,72.0bp) [draw,rectangle] {Linear | tanh};
  \node (t) at (260.0bp,18.0bp) [draw,rectangle] {Linear};
  \node (x) at (10.0bp,45.0bp) [draw,draw=none] {x};
  \node (l1) at (60.0bp,45.0bp) [draw,rectangle] {Linear | ReLU};
  \node (l2) at (140.0bp,45.0bp) [draw,rectangle] {Linear | ReLU};
  \node (s_out) at (320.0bp,72.0bp) [draw,draw=none] {$\a$};
  \node (t_out) at (320.0bp,18.0bp) [draw,draw=none] {$\b$};
  \draw [->] (x) -- (l1);
  \draw [->] (l1) -- (l2);
  \draw [->] (l2) -- (s);
  \draw [->] (l2) -- (t);
  \draw [->] (s) -- (s_out);
  \draw [->] (t) -- (t_out);
\end{tikzpicture}
    \caption{Illustration of coupling layer layout.}
    \label{fig:coupling_layer}
\end{figure}
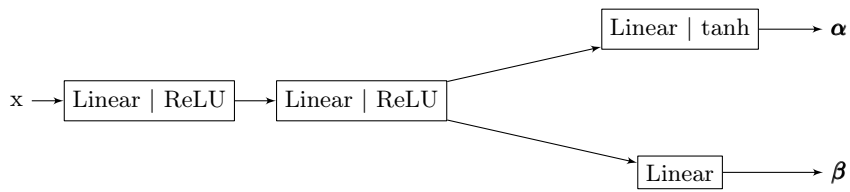

For our experiments we construct a normalizing flow model from affine coupling layers.
We use a model very similar to the one used by Kirichenko \textit{et al}. \cite{kirichenko}.
Our model consists of a series of 10 affine coupling layers,
alternating between transforming the first and second half of the input vector.
Each coupling layer consists of two shared linear layers with 512 hidden units followed by two parallel linear layers,
one for $\a$ and one for $\b$.
This is illustrated in Figure \ref{fig:coupling_layer}.

We train our model on features extracted from a YOLOv4 \cite{YOLOv4} object detection network.
Training is done for 200 epochs with a batch size of 128, using the Adam \cite{Adam} optimizer with a learning rate of $10^{-4}$.

\subsection{Intel RealSense D435}
Our experiments use an Intel RealSense D435\cite{D435}, which is a RGB+Depth camera.
In our experiments, however, we only utilise the RGB sensor.
Like most commercially available cameras, the D435 exposes a number of parameters that affect the resulting image in various ways.
For our experiments we manipulate a selection of these parameters, specifically:
backlight compensation, brightness, contrast, exposure, gain, saturation, and sharpness.

\subsection{Training Normalizing Flows on a Large Diverse Dataset} \label{sec:coco_flow}
In this section, we train a normalizing flow model using the entire COCO training dataset as our in-distribution.
We then test the model by comparing the log-likelihood outputs from different input image distributions.
Specifically, we divide this section into two smaller experiments:
\begin{itemize}
    \item One where we compare log-likelihoods of COCO images to randomly generated images.
    \item One where we capture many images of a static scene with randomized camera parameters,
          and compare their log-likelihood to COCO images.
\end{itemize}

\subsubsection{For our first experiment}
\begin{figure}[h!]
    \centering
    \includegraphics[width=\textwidth]{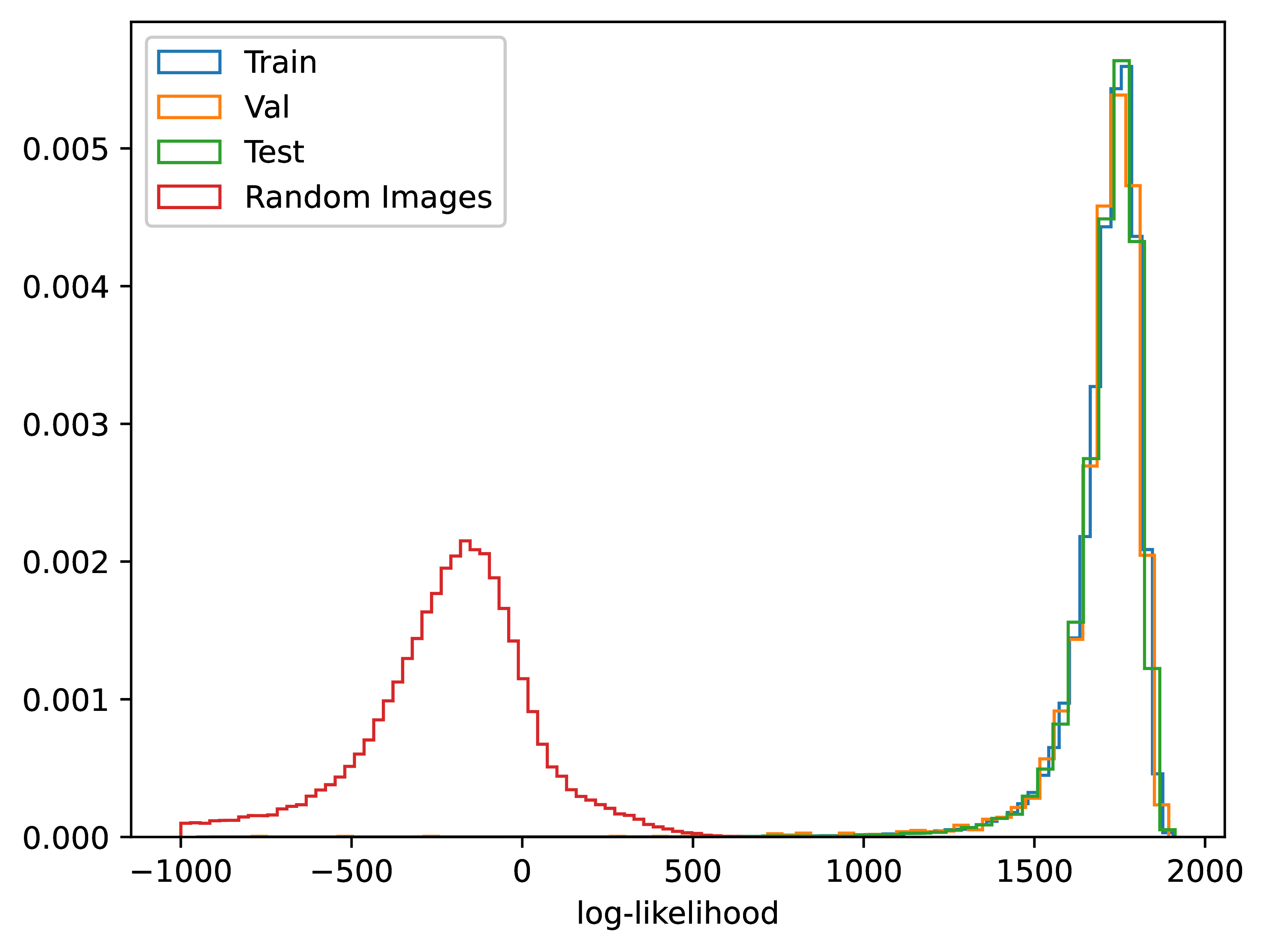}
    \caption{
        Normalized histogram plot of log-likelihoods from COCO training, validation and test datasets, as well as random images.
        The histogram has been cut at -1000 log-likelihood for improved readability.
    }
    \label{fig:nvp_yolo_hist}
\end{figure}

we run the trained model on all images in the COCO training, validation and testing datasets,
and record the resulting log-likelihood scores.
In order to show that the model can distinguish images that are \textit{definitely not} part of the in-distribution,
we also record log-likelihood scores from randomly generated images.
These random images were generated according to:
\begin{align*}
    \sigma &\sim \operatorname{Uniform}(1, 256) \\
    v_{i,j} &\sim \N(0, \sigma) \\
    pixel_{i,j} &= \min(255, \abs{v_{i,j}}), \ \ \forall \ i, j .
\end{align*}
Our primary motivation behind using this formula is to generate images with varying pixel ranges,
while also generating some mostly-black and mostly-white images.
By varying $\sigma$ per image, we vary the pixel range in different images.
When $\sigma$ is small, we get mostly-black images, and when $\sigma$ is large we get mostly-white images due to the $\min$ operation.
Figure \ref{fig:nvp_yolo_hist} shows histograms of randomly generated images,
along with the COCO training, validation, and testing datasets.

\subsubsection{For our second experiment}
\begin{figure}[h!]
    \centering
    \includegraphics[width=\textwidth]{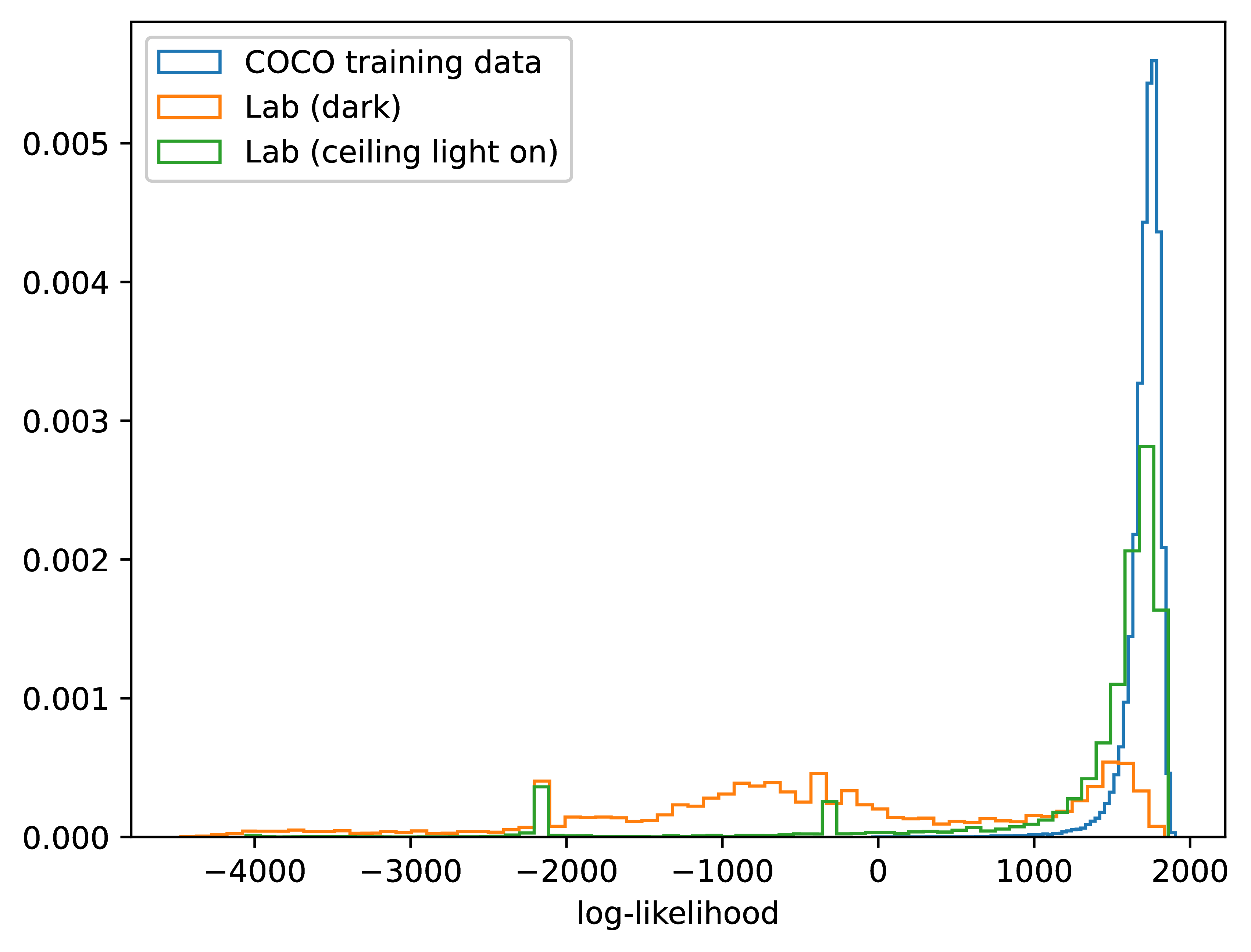}
    \caption{
        Normalized histogram plot of log-likelihood values.
        Blue: COCO training images.
        Orange: 10000 images in our lab with randomized camera parameters, without any lights on.
        Green: 10000 images in our lab with randomized camera parameters, with ceiling lights on.
    }
    \label{fig:nvp_coco_d435}
\end{figure}

we explore whether our normalizing flow can distinguish between different real images.
We place a stationary D435 camera in our lab, and capture a large number of images.
For each image, all camera parameters are randomized according to a uniform distribution.
In order to be certain that all variance in log-likelihood scores are caused by changing camera parameters,
we take precautions to minimize the amount of natural light entering our lab.
Then, we capture 10000 images with the ceiling lamp on, and 10000 with the ceiling lamp off.
The number 10000 was chosen arbitrarily.
Figure \ref{fig:nvp_coco_d435} shows the corresponding histograms of log-likelihood scores,
along with the COCO training data for comparison.

\subsection{Parameter Optimization} \label{sec:experiment}
Here, we present results from a small-scale experiment conducted in one of our robot labs.
In order to make the experiment as realistic as possible, the experiment was set up using one of our robots.
Our robot has an industrial arm with an Intel Realsense D435 camera attached at the gripper.

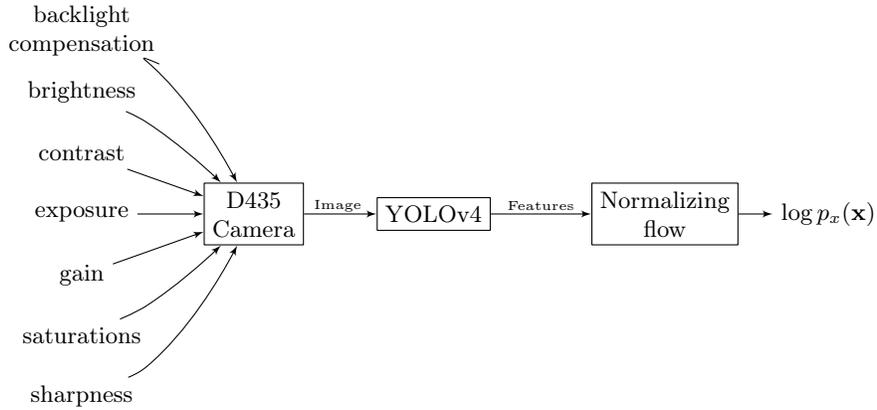
\begin{figure}[h!]
    \begin{tikzpicture}[>=latex',line join=bevel,scale=0.425]
\begin{scope}
  \pgfsetstrokecolor{black}
  \definecolor{strokecol}{rgb}{1.0,1.0,1.0};
  \pgfsetstrokecolor{strokecol}
  \definecolor{fillcol}{rgb}{1.0,1.0,1.0};
  \pgfsetfillcolor{fillcol}
  \filldraw (0.0bp,0.0bp) -- (0.0bp,362.0bp) -- (776.0bp,362.0bp) -- (776.0bp,0.0bp) -- cycle;
\end{scope}
  \node (bc) at (58.0bp,343.0bp) [draw,draw=none,align=center] {backlight \\ compensation};
  \node (br) at (58.0bp,288.0bp) [draw,draw=none] {brightness};
  \node (ct) at (58.0bp,234.0bp) [draw,draw=none] {contrast};
  \node (ex) at (58.0bp,180.0bp) [draw,draw=none] {exposure};
  \node (ga) at (58.0bp,126.0bp) [draw,draw=none] {gain};
  \node (st) at (58.0bp,72.0bp) [draw,draw=none] {saturations};
  \node (sp) at (58.0bp,18.0bp) [draw,draw=none] {sharpness};
  \node (camera) at (210.5bp,180.0bp) [draw,rectangle,align=center] {D435 \\ Camera};
  \node (yolo) at (370.0bp,180.0bp) [draw,rectangle] {YOLOv4};
  \node (nf) at (575.0bp,180.0bp) [draw,rectangle,align=center] {Normalizing \\ flow};
  \node (score) at (720.0bp,180.0bp) [draw,draw=none] {$\log p_x(\mathbf{x})$};
  \draw [->] (bc) ..controls (108.04bp,321.2bp) and (112.24bp,318.24bp)  .. (116.0bp,315.0bp) .. controls (151.68bp,284.28bp) and (180.23bp,237.02bp)  .. (camera);
  \draw [->] (br) ..controls (105.85bp,267.19bp) and (111.14bp,264.17bp)  .. (116.0bp,261.0bp) .. controls (141.02bp,244.67bp) and (166.46bp,222.17bp)  .. (camera);
  \draw [->] (ct) ..controls (112.21bp,214.91bp) and (131.34bp,208.04bp)  .. (camera);
  \draw [->] (ex) ..controls (112.7bp,180.0bp) and (127.96bp,180.0bp)  .. (camera);
  \draw [->] (ga) ..controls (99.477bp,140.52bp) and (125.46bp,149.85bp)  .. (camera);
  \draw [->] (st) ..controls (105.85bp,92.813bp) and (111.14bp,95.829bp)  .. (116.0bp,99.0bp) .. controls (141.02bp,115.33bp) and (166.46bp,137.83bp)  .. (camera);
  \draw [->] (sp) ..controls (107.3bp,38.396bp) and (111.91bp,41.528bp)  .. (116.0bp,45.0bp) .. controls (151.89bp,75.462bp) and (180.39bp,122.8bp)  .. (camera);
  \draw [->] (camera) ..controls (291.07bp,180.0bp) and (316.97bp,180.0bp)  .. (yolo);
  \definecolor{strokecol}{rgb}{0.0,0.0,0.0};
  \pgfsetstrokecolor{strokecol}
  \draw (285.0bp,187.5bp) node {\tiny Image};
  \draw [->] (yolo) ..controls (448.4bp,180.0bp) and (482.95bp,180.0bp)  .. (nf);
  \draw (465.0bp,187.5bp) node {\tiny Features};
  \draw [->] (nf) ..controls (635.75bp,180.0bp) and (644.69bp,180.0bp)  .. (score);
\end{tikzpicture}
    \caption{Parameter optimization setup.}
    \label{fig:opt_setup}
\end{figure}

For this experiment, we set up an optimization procedure using camera parameters as our input,
and the log-likelihood from our normalizing flow model as output, as illustrated in Figure \ref{fig:opt_setup}.
More formally, let $\theta$ denote the camera parameters, and $C(\theta)$ denote the camera
(a function that takes camera parameters and produces an image $\x$), then we solve:
\[
    \argmax_{\theta} \ \log p_x(C(\theta))
\]
\[
    =
    \argmax_{\theta} \ \log p_u(T^{-1}(C(\theta))) + \log \abs{\det J_{T^{-1}}(C(\theta))}.
\]

For the optimization procedure, we used a very simple evolutionary optimization with elitism selection,
a population of 50, and a mutation rate of 20\%.
No crossover was used.
Optimization was carried out for a total of 200 evaluations.
Population size, mutation rate, and number of evaluations were chosen based on prior experience with evolutionary optimization.
A small number of trial runs confirmed that these values work well.

In our lab, we set up 13 different scenarios with varying COCO objects, viewing angles, and lighting conditions.
We took care to make the scenarios as challenging as possible by introducing difficult lighting conditions,
reflective surfaces, and background clutter.
With the robot stationary, optimization was performed as described above.
During optimization, the images with the highest and lowest log-likelihood scores were saved along with their corresponding parameters,
and compared to images captured with the camera's default parameter settings.
Henceforth, these will be referred to as \textit{best}, \textit{worst} and \textit{default} parameters, respectively.

After the optimization procedure, all images were annotated with bounding boxes for the COCO objects present.
They were then fed through YOLOv4.
Mean-Average-Precision (mAP), mean-Average-Recall (mAR), and F1 scores were computed and compared between the
worst, default, and best camera parameters.

Table \ref{tbl:average_scores} shows average mAP, mAR, and F1 scores for best, default and worst camera parameters.
In table \ref{tbl:best_default_gain} we show statistics for the improvement in
mAP, mAR, and F1 scores of the best parameters compared to the default parameters.
Similarly, Table \ref{tbl:best_worst_gain} shows statistics for the improvement
of the best compared to the worst parameters.

\begin{table}[h!]
    \centering
    \caption{Average mAP, mAR, and F1 score for best, default, and worst parameters.}
    \begin{tabular}{|r|l|l|l|}
        \hline
        & \textbf{mAP} & \textbf{mAR} & \textbf{F1} \\
        \hline
        \textbf{Best} & 0.7977 & 0.3283 & 0.4605 \\
        \hline
        \textbf{Default} & 0.7591 & 0.2977 & 0.4219 \\
        \hline
        \textbf{Worst} & 0.1731 & 0.0431 & 0.0687 \\
        \hline
    \end{tabular}
    \label{tbl:average_scores}
\end{table}

\begin{table}[h!]
    \centering
    \caption{mAP, mAR, and F1 improvement. Best compared to default parameters.}
    \begin{tabular}{|r|l|l|l|}
        \hline
        & \textbf{Min} & \textbf{Max} & \textbf{Average} \\
        \hline
        \textbf{mAP} & -0.2889 & +0.5778 & +0.0386 \\
        \hline
        \textbf{mAR} & -0.1083 & +0.25 & +0.0306 \\
        \hline
        \textbf{F1} & -0.1056 & +0.349 & +0.0386 \\
        \hline
    \end{tabular}
    \label{tbl:best_default_gain}
\end{table}

\begin{table}[h!]
    \centering
    \caption{mAP, mAR, and F1 improvement. Best compared to worst parameters.}
    \begin{tabular}{|r|l|l|l|}
        \hline
        & \textbf{Min} & \textbf{Max} & \textbf{Average} \\
        \hline
        \textbf{mAP} & -0.104 & +0.95 & +0.6247 \\
        \hline
        \textbf{mAR} & +0.1375 & +0.4278 & +0.2852 \\
        \hline
        \textbf{F1} & +0.1794 & +0.5703 & +0.3918 \\
        \hline
    \end{tabular}
    \label{tbl:best_worst_gain}
\end{table}

\begin{figure}[h!]
    \includegraphics[width=0.49\textwidth]{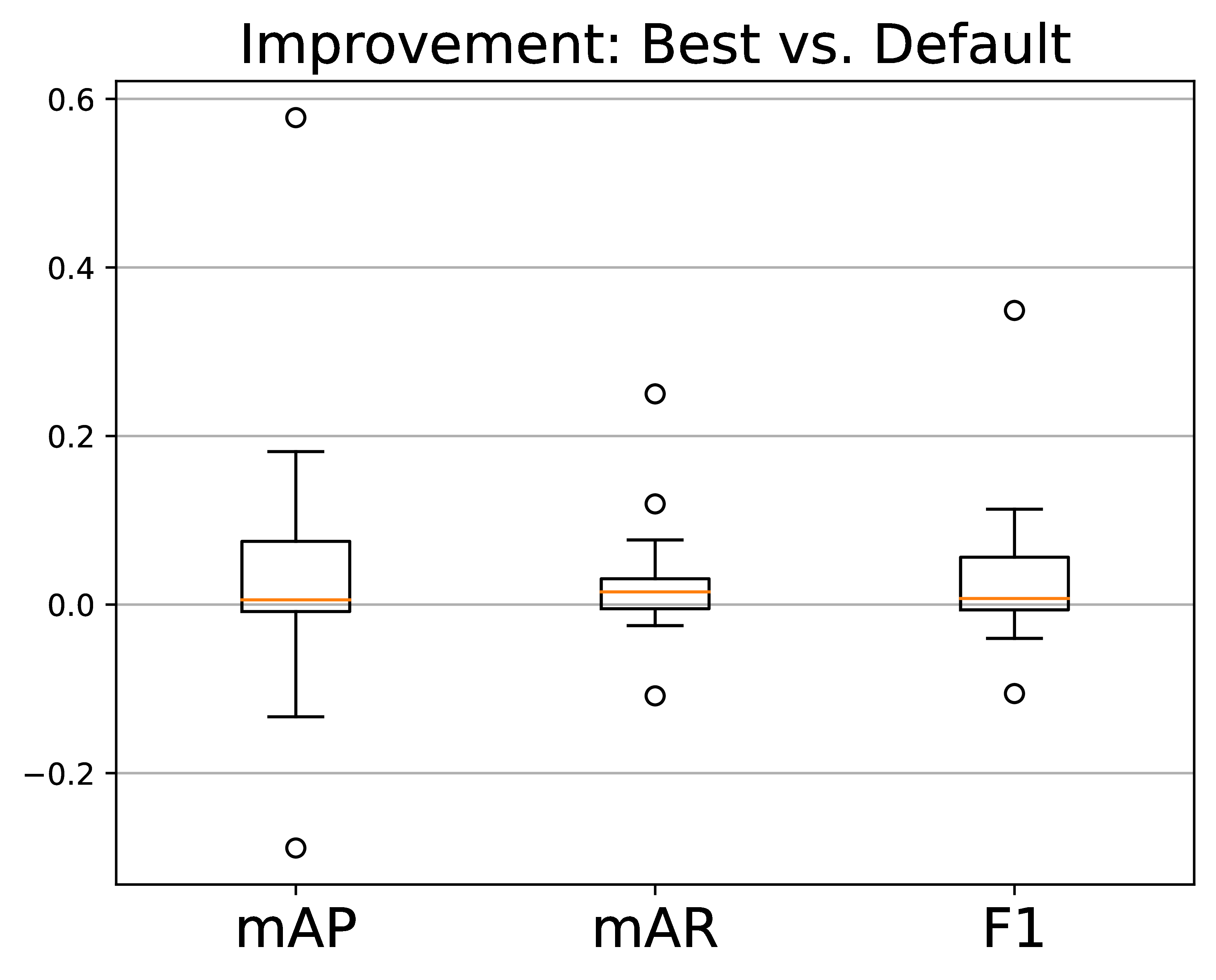}
    \includegraphics[width=0.49\textwidth]{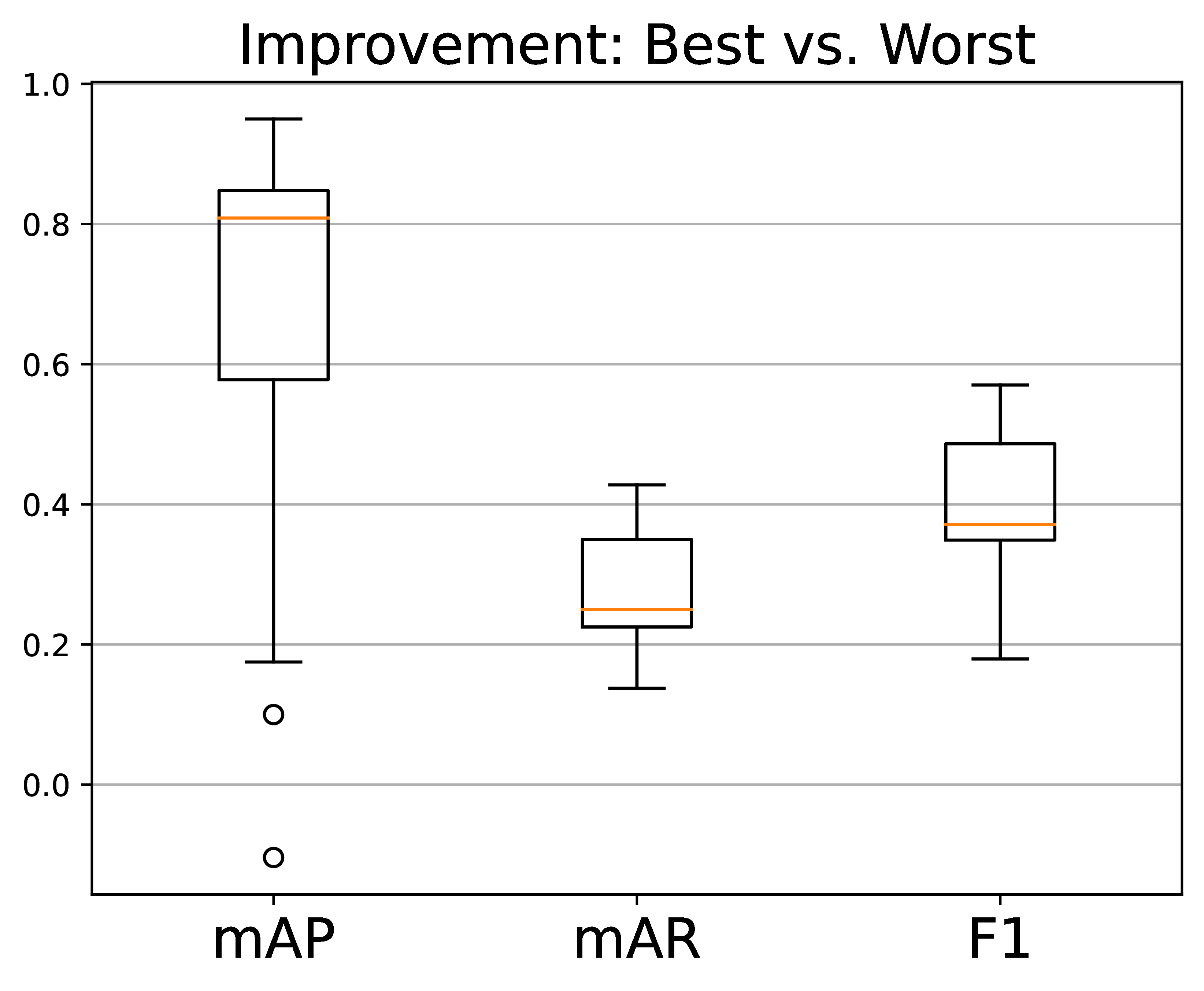}
    \caption{
        Boxplot of mAP, mAR, and F1 improvement.
        Left: Best compared to default parameters.
        Right: Best compared to worst parameters.
    }
    \label{fig:best_gain}
\end{figure}

Figure \ref{fig:best_gain} shows box plots of the improvements in
mAP, mAR, and F1 scores for the best parameters compared with the
default and worst parameters respectively.

\begin{figure}[h!]
    \centering
    \begin{tabular}{ccc}
        \textbf{Worst} & \textbf{Default} & \textbf{Best} \\
        \includegraphics[width=0.3\textwidth]{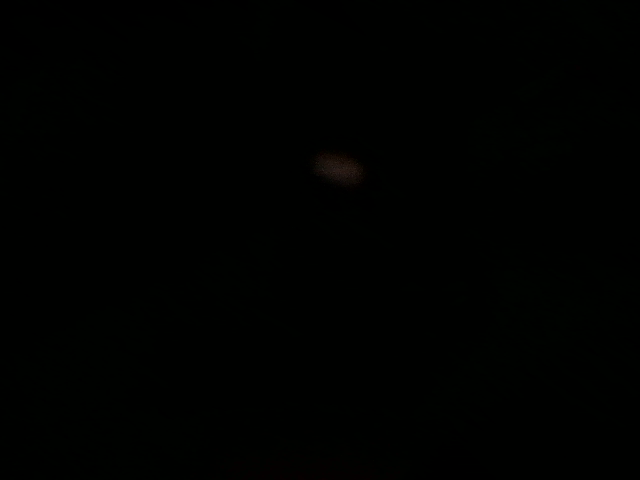} &
        \includegraphics[width=0.3\textwidth]{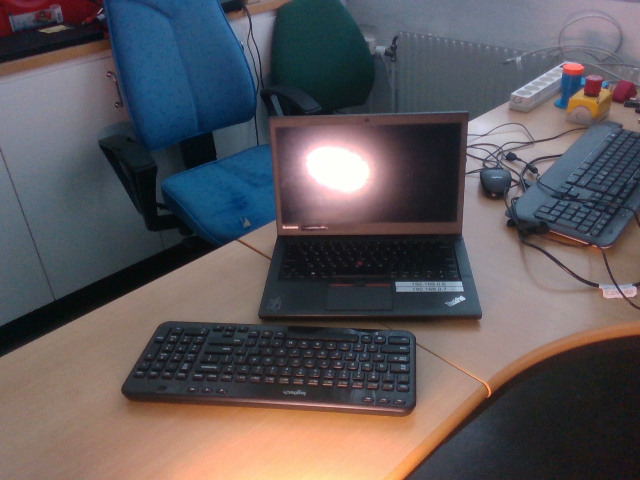} &
        \includegraphics[width=0.3\textwidth]{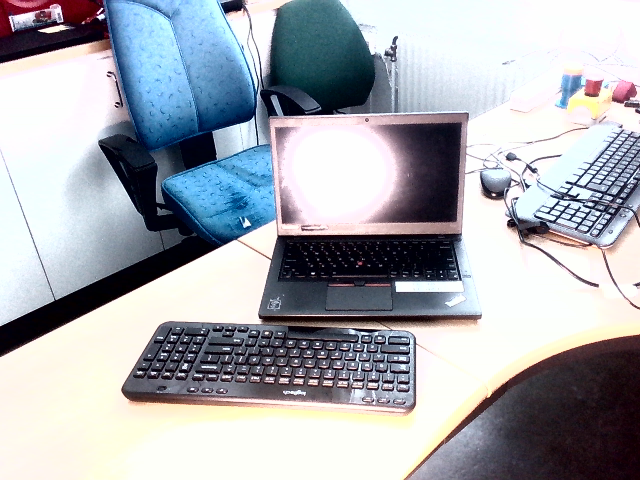} \\
        \vspace{2ex}
        F1: 0.0 & F1: 0.4464 & F1: 0.5595 \\
        \includegraphics[width=0.3\textwidth]{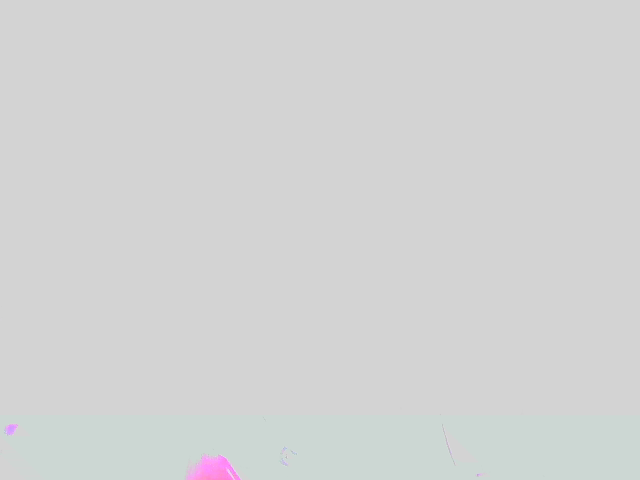} &
        \includegraphics[width=0.3\textwidth]{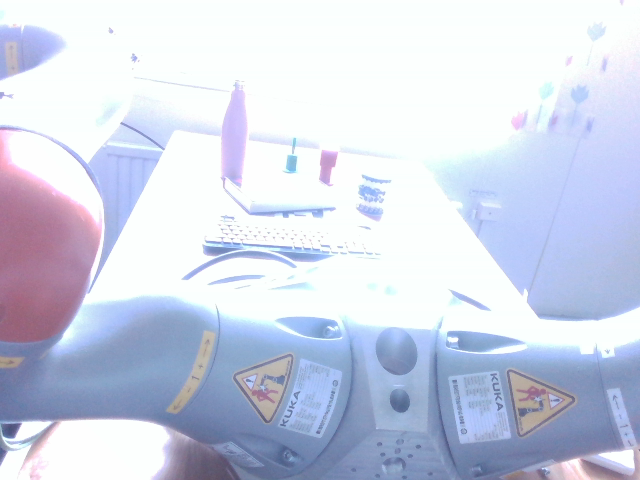} &
        \includegraphics[width=0.3\textwidth]{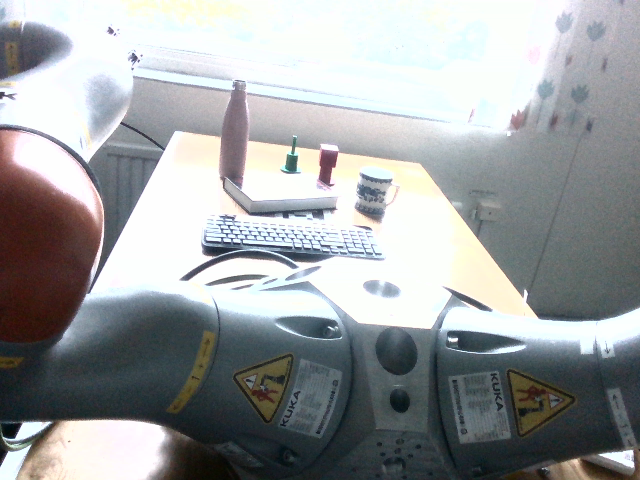} \\
        F1: 0.0 & F1: 0.0 & F1: 0.3490
    \end{tabular}
    \caption{
        Examples where parameter optimization resulted in an improvement in object detection.
        Left: lowest log-likelihood.
        Center: default parameters.
        Right: highest log-likelihood.
    }
    \label{fig:good_examples}
\end{figure}

\begin{figure}[h!]
    \centering
    \begin{tabular}{ccc}
        \textbf{Worst} & \textbf{Default} & \textbf{Best} \\
        \includegraphics[width=0.3\textwidth]{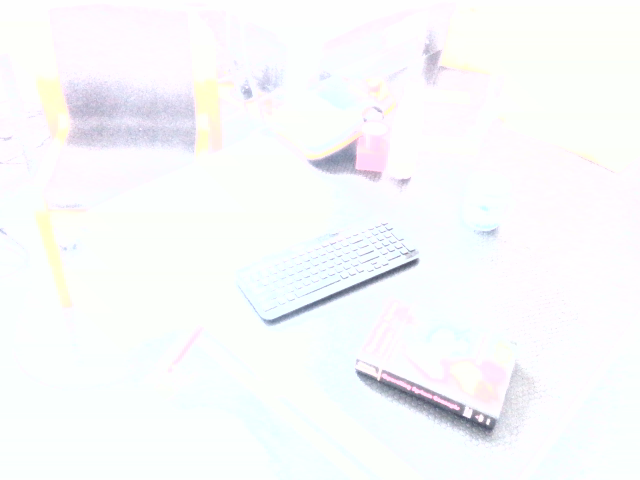} &
        \includegraphics[width=0.3\textwidth]{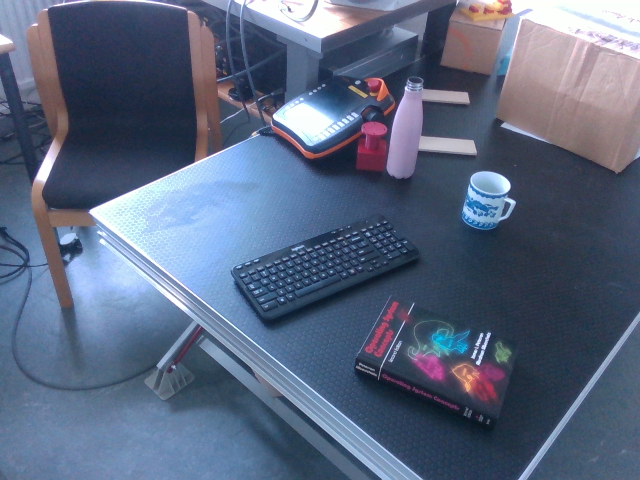} &
        \includegraphics[width=0.3\textwidth]{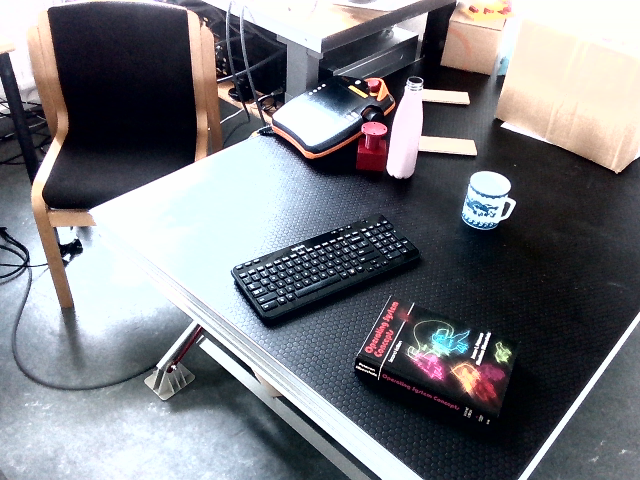} \\
        \vspace{2ex}
        F1: 0.3864 & F1: 0.6127 & F1: 0.6289 \\
        \includegraphics[width=0.3\textwidth]{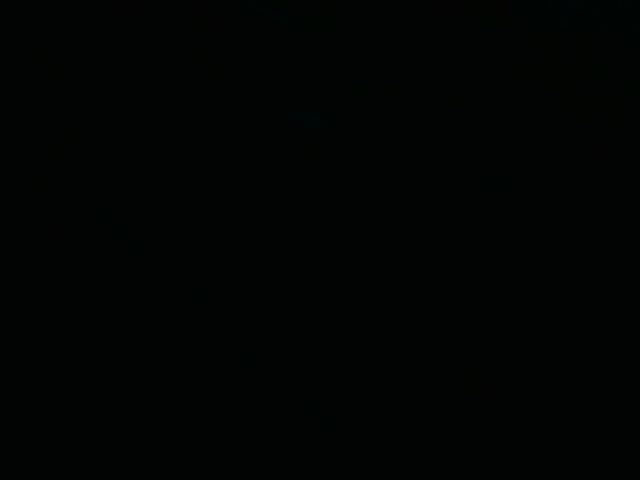} &
        \includegraphics[width=0.3\textwidth]{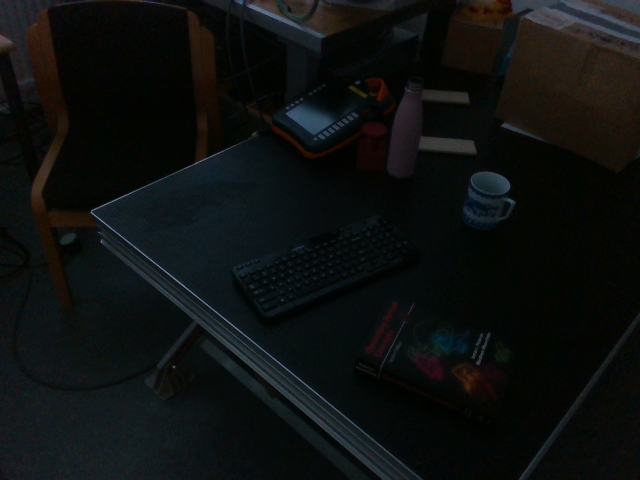} &
        \includegraphics[width=0.3\textwidth]{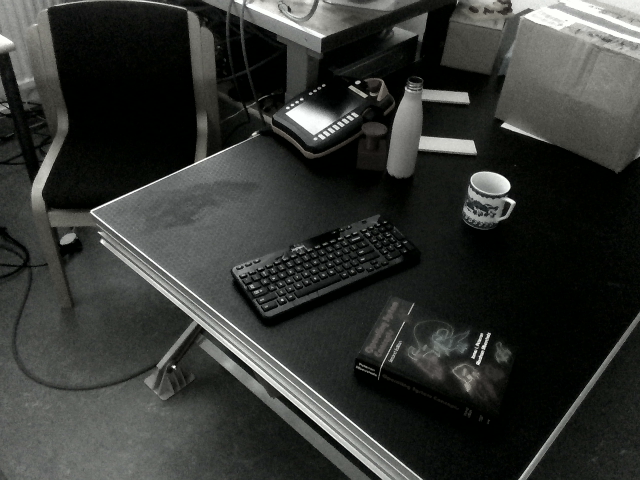} \\
        F1: 0.0 & F1: 0.4793 & F1: 0.4865
    \end{tabular}
    \caption{
        Examples where parameter optimization resulted in no significant difference in object detection.
        Left: lowest log-likelihood.
        Center: default parameters.
        Right: highest log-likelihood.
    }
    \label{fig:negligible_examples}
\end{figure}

\begin{figure}[h!]
    \centering
    \begin{tabular}{ccc}
        \textbf{Worst} & \textbf{Default} & \textbf{Best} \\
        \includegraphics[width=0.3\textwidth]{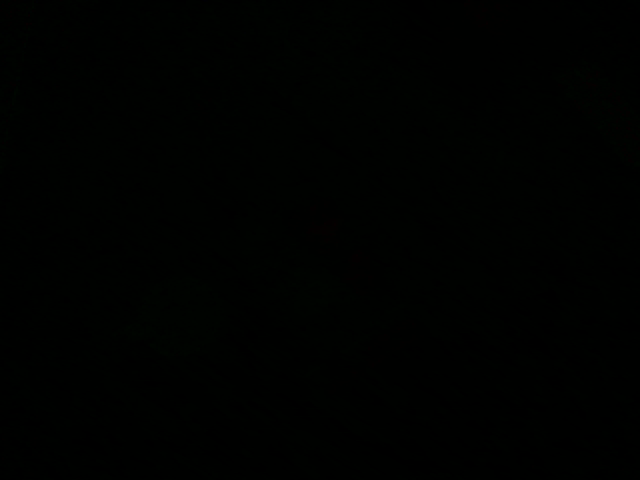} &
        \includegraphics[width=0.3\textwidth]{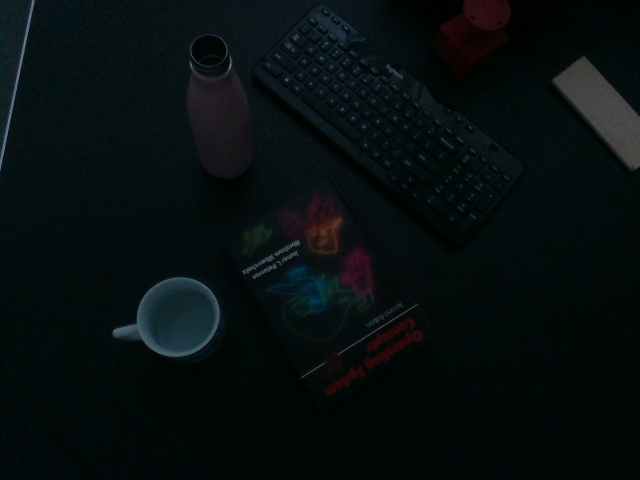} &
        \includegraphics[width=0.3\textwidth]{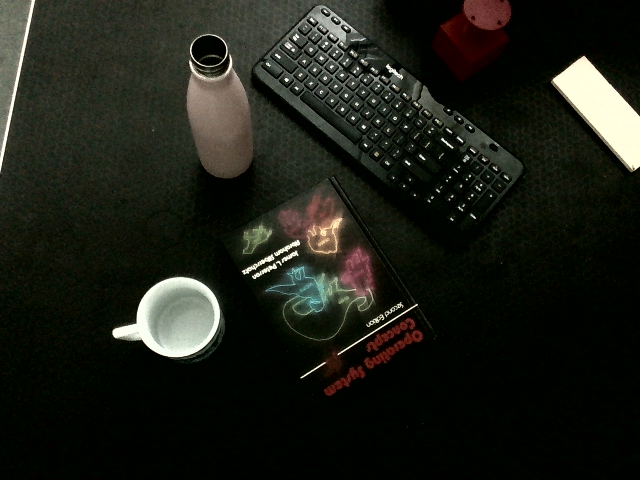} \\
        F1: 0.0 & F1: 0.6172 & F1: 0.5115
    \end{tabular}
    \caption{
        Example where parameter optimization resulted in worse performance in object detection.
        Left: lowest log-likelihood.
        Center: default parameters.
        Right: highest log-likelihood.
    }
    \label{fig:bad_examples}
\end{figure}

In Figures \ref{fig:good_examples}, \ref{fig:negligible_examples}, and \ref{fig:bad_examples} we show example scenarios where optimization led to varying degrees of success.
Each figure includes images along with the respective F1 scores from YOLOv4.
Figure \ref{fig:good_examples} shows examples where parameter optimization resulted in
better object detection than the default parameters.
Figure \ref{fig:negligible_examples} shows examples where optimiztion resulted in no significant difference in object detection.
Finally, Figure \ref{fig:bad_examples} shows the only scenario where optimization resulted in significantly worse object detection.

\section{Discussion}

\subsection{Training on the COCO Dataset}
When training on the full COCO dataset we see two important indicators that the model is learning properly.
First, as seen in Figure \ref{fig:nvp_yolo_hist}, the output log-likelihoods span a wide range of values.
It would be a red flag if all log-likelihoods were too similar, indicating that the learned probability distribution might be flat.
Second, we observe that the validation and test datasets are distributed the same way as the training set,
while the random images are given distinctly lower log-likelihood scores.

In figure \ref{fig:nvp_coco_d435}, we see that the model is able to distinguish between real images from different distributions.
As such, we have found no evidence to suggest that the size and diversity of the COCO dataset should pose any problems for OOD detection.
This in turn suggests that normalizing flow models can be readily applied to diverse real-world scenarios.

Additionally, figure \ref{fig:nvp_coco_d435} shows that varying parameters in the D435 over the same
scene can result in a wide range of log-likelihood values.
This gives merit to the idea that camera parameters form a good basis for adapting to various imaging conditions.

\subsection{Parameter Optimization Experiment}
Table \ref{tbl:average_scores} tells us at a glance
that our optimized camera parameters acheived the best average scores across the board.
Looking at Table \ref{tbl:best_default_gain}, we see an average improvement of 3 to 4 percentage points in terms of all metric scores.
From the box plot in Figure \ref{fig:best_gain},
we see that most scenarios yield only a small difference in object detection compared to the default parameters.
However, most importantly we see that the trend is a positive improvement.
We also observe that the positive outliers are significantly larger than the negative outliers, again suggesting a positive trend.

Perhaps unsurprisingly, Table \ref{tbl:best_worst_gain} shows a massive improvement (28 to 62 percentage points on average) between the best and worst parameters found.
The box plot in Figure \ref{fig:best_gain} further highlights this improvement.

In Figures \ref{fig:good_examples}, \ref{fig:negligible_examples}, and \ref{fig:bad_examples} we note that
some of the parameter-optimized pictures look better to the human eye,
while others look worse despite resulting in better object detection.
This highlights the important fact that there is no equivalence between images that look good to the human eye,
and images that are good for object detection.

Perhaps the most important result of this experiment is not the fact that parameter-optimization can lead to better object detection,
but instead that it seems feasible to adapt and recover from a very poor visual scenario.
Extrapolating from the fact that different camera parameters could result in images where YOLO performs well,
and where it fails almost completely on the same scenario,
it seems feasible that these parameters have the necessary capacity to adapt to a wide range of difficult visual conditions.
This is given further merit by the fact that the varying camera parameters resulted in widely varying log-likelihood values in
figure \ref{fig:nvp_coco_d435}.
As such, when the normalizing flow model detects something with extremely low log-likelihood value,
it is most likely possible to adapt camera parameters to get an image more suitable for object detection. 
While these experiments have only been concerned with object detection,
we expect that this behaviour generalises to other vision tasks as well.

\section{Conclusion}
In this paper we have shown that it is possible to train a normalizing flow model
for OOD detection on a large and diverse dataset such as COCO.
We have also conducted a real-world experiment which shows that a normalizing
flow OOD detector can be used as a quality metric for a vision task.
Optimizing camera parameters with respect to the output from our normalizing flow OOD model yielded, on average,
a 3 to 4 percent unit improvement in the reliability of YOLOv4 in terms of object detection performance.

\section{Future Work}
This paper lays the groundwork for a larger framework of quality metrics for vision task.
In order to further expand this framework, a larger benchmark dataset is beneficial.
It is also an interesting research direction to explore whether image improvements can be \textit{learned} based on this OOD metric,
rather than going through the somewhat slow optimization process used in our experiments.

\subsubsection{Acknowledgements}
This research is funded by the Excellence Center at Linköping-Lund
in Information Technology (ELLIIT), and the Wallenberg AI, Autonomous Systems
and Software Program (WASP). Computations for this publication were enabled
by the supercomputing resource Berzelius provided by the National Supercomputer
Centre at Linköping University and the Knut and Alice Wallenberg foundation.
Additionally, Luigi Nardi was supported in part by affiliate members and other supporters
of the Stanford DAWN project – Ant Financial, Facebook, Google, Intel, Microsoft,
NEC, SAP, Teradata, and VMware.

%
%

\bibliographystyle{splncs04}
\bibliography{paper}

\end{document}